\documentclass[letterpaper, 10 pt, conference]{ieeeconf}  % Comment this line out if you need a4paper
\IEEEoverridecommandlockouts                              % This command is only needed if 
\usepackage{comment}
\usepackage{amsmath}
\usepackage{mathtools}
\usepackage{algorithm}
\usepackage{array}
\usepackage[noend]{algpseudocode}
\usepackage{amssymb} 
\usepackage{graphicx}
\usepackage[font=footnotesize]{subcaption}
\usepackage[font=footnotesize]{caption}
\usepackage{hyperref}
\usepackage{amssymb}
\usepackage{url}
\usepackage{color}
\usepackage{wrapfig}
\usepackage{xcolor,colortbl}
\usepackage{ifthen}
\usepackage{xspace}
\usepackage{multirow}
\usepackage[normalem]{ulem}
\usepackage{wrapfig}
\renewcommand{\baselinestretch}{0.975}

\title{\LARGE \bf Planning for Multi-stage Forceful Manipulation}

\author{Rachel Holladay$^{1}$, Tom\'{a}s Lozano-P\'{e}rez$^{1}$, and Alberto Rodriguez$^{2}$% <-this % stops a space
\thanks{$^{1}$Rachel Holladay and Tom\'{a}s Lozano-P\'{e}rez are with the Computer Science and Artificial Intelligence Laboratory, Massachusetts Institute of Technology, {\tt\small \{rhollada,tlp@mit\}.edu}.}
\thanks{$^{2}$Alberto Rodriguez is with the Mechanical Engineering Department, Massachusetts Institute of Technology, {\tt\small albertor@mit.edu}.}
}

\begin{document}

\maketitle
\thispagestyle{empty}
\pagestyle{empty}

% Labels in IEEE format
\newcommand{\eref}[1]{(\ref{#1})} % Equation
\newcommand{\sref}[1]{Sec. \ref{#1}} % Section
\newcommand{\figref}[1]{Fig.\ref{#1}} % Figure
\newcommand{\tref}[1]{Table \ref{#1}} %Table
\newcommand{\aref}[1]{Algorithm \ref{#1}} %Algorithm
\newcommand{\lref}[1]{Line \ref{#1}} %Line in Algorithm
\renewcommand*\rmdefault{ppl}
\setlength{\textfloatsep}{5pt}

\newcommand{\etal}{et al.}

\newboolean{include-notes}
\setboolean{include-notes}{true}
\newcommand{\rhnote}[1]{\ifthenelse{\boolean{include-notes}}%
 {\textcolor{red}{\textbf{RH: #1}}}{}}
\newcommand{\rhtodo}[1]{\ifthenelse{\boolean{include-notes}}%
 {\textcolor{orange}{\textbf{RH DO: #1}}}{}}
\newcommand{\tlpnote}[1]{\ifthenelse{\boolean{include-notes}}%
 {\textcolor{blue}{\textbf{TLP: #1}}}{}}
\newcommand{\arnote}[1]{\ifthenelse{\boolean{include-notes}}%
 {\textcolor{brown}{\textbf{AR: #1}}}{}}

\begin{abstract}
Multi-stage forceful manipulation tasks, such as twisting a nut on a bolt, require reasoning over interlocking constraints over discrete and continuous choices.
The robot must choose a sequence of discrete actions, or strategy, such as whether to pick up an object, and the continuous parameters of each of those actions, such as how to grasp that object. 
In forceful manipulation tasks, the force requirements substantially impact the choices of both strategy and parameters.
To enable planning and executing forceful manipulation, we augment an existing task and motion planner with controllers that exert wrenches and constraints that explicitly consider torque and frictional limits. 
In two domains, opening a childproof bottle and twisting a nut, we demonstrate how the system considers a combinatorial number of strategies and how choosing actions that are robust to parameter variations impacts the choice of strategy. \href{https://mcube.mit.edu/forceful-manipulation/}{https://mcube.mit.edu/forceful-manipulation/}
\end{abstract}

% !TEX root = main.tex

\section{Introduction}
\label{sec:intro}

%1. What is the problem?   
%2. Why is it relevant?
%3. Why is it hard?
%4. What have others done?
%5. What's missing?
%6. What is our ONE key insight?           
%7. How do we compare against the state of the art?
%8. What are our contributions?
%9. What are our limitations?

%%%%%%%%%%%%%%%%%%%%%%%%%%%%%%%%%%%%%%%%%%%%%%%%%%%%%%%%%%%%%%%%%%%%%%%%%%%%%%%
%\tlpnote{tomas} \arnote{alberto}

Our goal is to enable robots to plan and execute {\em forceful manipulation tasks} such as drilling through a board, cutting a carrot, and twisting a nut.  
While all tasks that involve contact are technically forceful, we refer to forceful manipulation tasks as those where the ability to generate and transmit the necessary forces to objects and their environment is an active limiting factor which must be considered. 
Respecting these limits might require a planner, for example, to prefer a more forceful kinematic configuration of the robot arm or a more stable grasp of an object. 

\begin{figure}[t]
\centering
    \includegraphics[width=0.48\columnwidth]{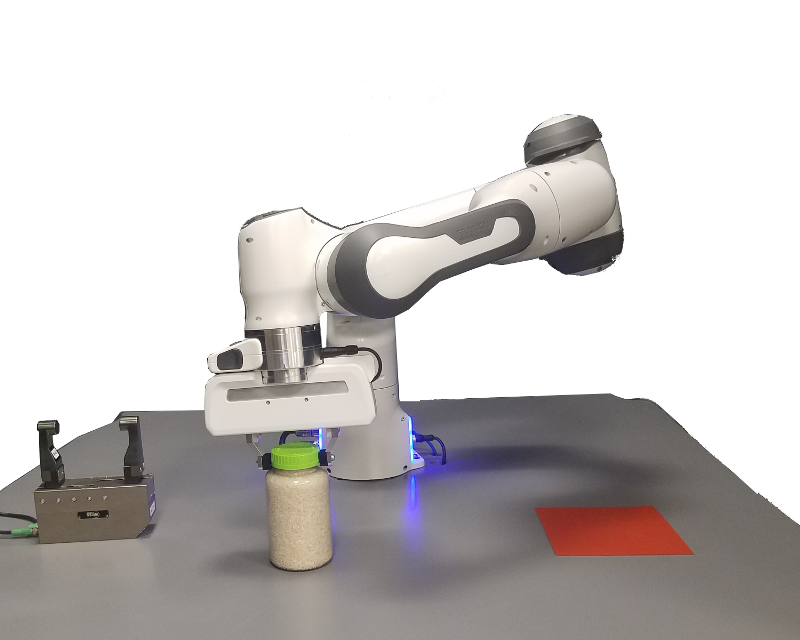}
    \includegraphics[width=0.48\columnwidth]{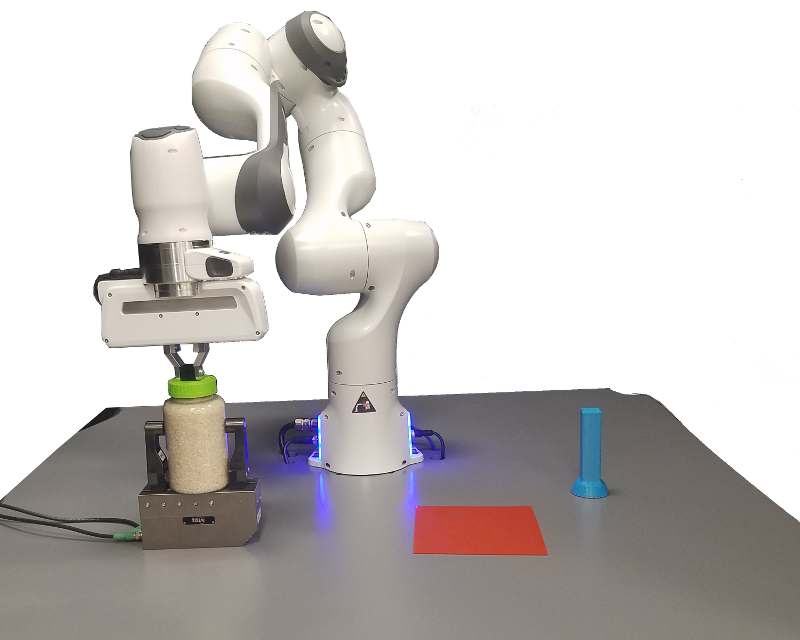}
    \includegraphics[width=0.4\columnwidth]{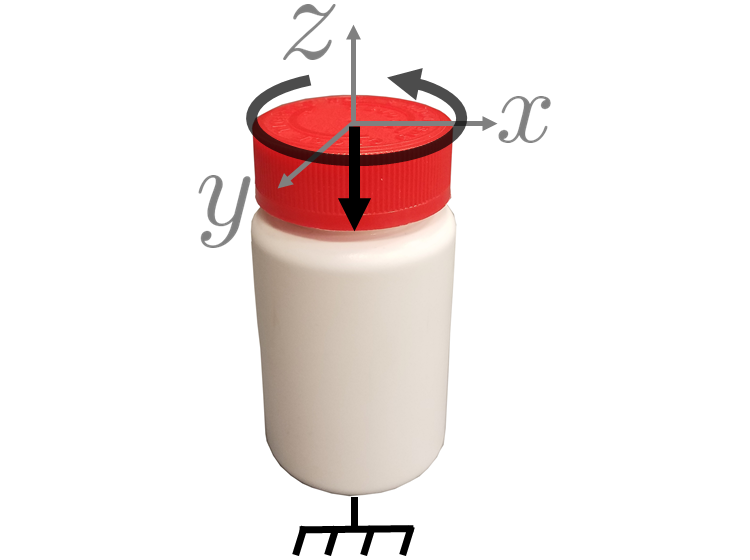}
    \includegraphics[width=0.4\columnwidth]{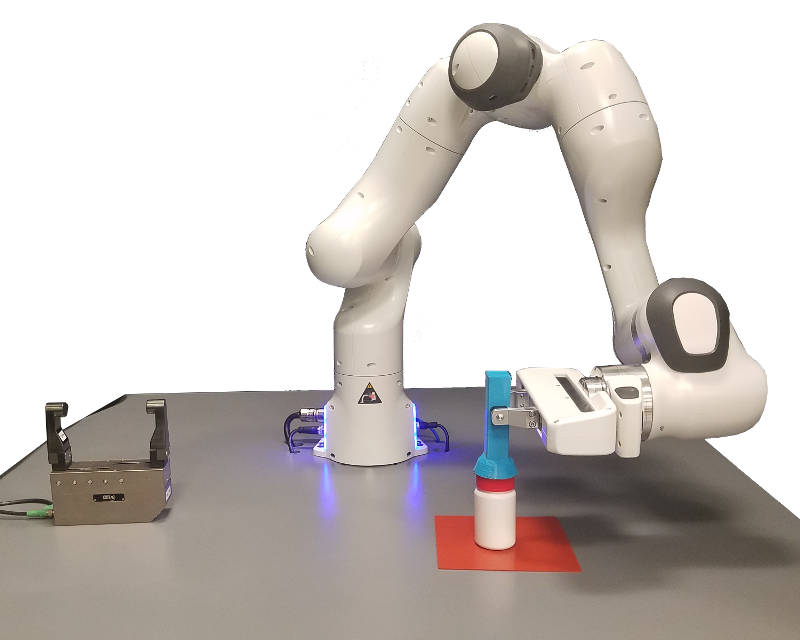}
\caption{Opening a childproof bottle involves executing a downward-push on the lid, while securing the bottle (lower left). Our system can reason over a combinatorial number of strategies to accomplish this forceful manipulation task, including twisting with various parts of its end effector, twisting with a tool (in blue), securing with a vise (in grey), securing against the table, or securing against a high-friction rubber mat (in red).}
\label{fig:fig1}
\end{figure}

Forceful operations, as defined by Chen \etal, are the exertion of a wrench (generalized force/torque) at a point on an object~\cite{chen2019manipulation}.  
These operations are intended to be quasi-statically stable, i.e. the forces are always in balance and produce relatively slow motions, and will generally require some form of fixturing to balance the applied wrenches.  
For example, opening a push-and-twist childproof bottle is an example of a forceful manipulation task. To remove the lid the robot must exert a downward force on the lid while applying a torque along the axis of the bottle (\figref{fig:fig1}-bottom left). 
The robot must be strong enough to deliver the required wrench and also secure the bottle before exerting a wrench on it, to insure that there is enough available friction between the fingers and the lid.

To accomplish complex, multi-step forceful manipulation tasks, robots need to make discrete decisions, such as whether to push on the lid with the fingers, the palm or a tool, and whether to secure the bottle via frictional contact with a surface, with another gripper or with a vise. 
The robot must also make continuous decisions such as the choice of grasp pose, initial robot configuration and object poses in the environment. 
Critically, all these decisions interact in relatively complex ways to achieve a valid task execution.

\figref{fig:fig1} illustrates that there are \textit{different strategies} for completing this task. 
Each strategy's viability depends on the robot's choices and on the environment. 
For example, a solution that uses the friction from the table to secure the bottle, as shown in the top left, would fail if the table can only provide a small amount of friction.  
%the friction coefficient between the table and the bottle were lower.  
Instead, the robot would need to find a significantly different strategy, such as securing, or fixturing, the bottle via a vise, as shown in the top right.

Strategies represent sequences of parameterized high-level actions.  Each action is implemented as  a controller parameterized by a set of constrained continuous values, such as robot configurations, grasp poses, trajectories, etc.  
%For example, a controller that exerts a forceful operation is, in part, parameterized by the applied wrench and the pose of the object. 
Our goal is to find both a sequence of high-level actions (a strategy) and parameter values for those actions, all of which satisfy the coupled constraints. 

To produce valid solutions for a wide range of object and environment configurations, the robot must be able to consider a wide range of strategies. 
%, where the validity of a strategy is determined by the existence of parameter values that satisfy the constraints imposed by the actions. 
As illustrated above, small changes, such as decreasing a friction coefficient, may necessitate an entirely new strategy. Approaches that attempt to explicitly encode solutions in the form of a policy, \textit{e.g.} via a finite state machine or a fixed action sequence, will generally fail to capture the full range of feasible strategies~\cite{michelman1994forming,holladay2019force}. 
Methods that attempt to learn such a policy will need a very large number of interactions to explore this rich and highly-constrained solution space. 

We propose addressing forceful manipulation problems by planning over a combinatorial set of discete/continuous strategies.  We extend an existing task and motion planning (TAMP) system, PDDLStream, which reduces this type of hybrid discrete/continuous planning problem to a sequence of discrete planning problems via focused sampling of the continuous parameters~\cite{garrett2020pddlstream}. 
To apply this method to forceful manipulation, we introduce new actions with controllers that exert wrenches and constraints that explicitly consider torque and frictional limits. 
In particular, the constraints propagate the wrench of the forceful operations throughout the actions in a strategy.
In the childproof bottle example, this propagation is critical to finding a strategy that fixtures the bottle against the wrench applied on the lid.
%Many of the constraints draw from fundamental tools in the mechanics of  manipulation, allowing us to exploit similarities in modeling structure.
%Additionally, some of the actions, in order to perform the forceful operations, involve controllers that exert wrenches on their environment. 

%With this approach, we can consider strategies that are combinatorial assemblies of the actions and choose a satisfying strategy based on the parameters of the environment. 
%We explore how the sampling-based planning method is impacted by the interlocking force constraints, both with respect to performance and to which strategies are found. 
%Furthermore, we demonstrate how to enable the planner to choose strategies that are robust, i.e. those that are not brittle with respect to the values of those parameters, by formulating this as cost-sensitive planning. 
%We formulate robust planning using cost-sensitive planning, where the cost of an action is tied to its probability of success in open-loop execution, given perturbations in parameters for the force-based constraints.

%In addition to opening a childproof lid, we also apply the system on the task of twisting a nut on a bolt. 
Our paper makes the following contributions: 
\begin{itemize}
    \item Augment an existing TAMP method~\cite{garrett2020pddlstream} for forceful manipulation tasks by adding controllers and constraints.
    \item Demonstrate, in two domains (opening a childproof bottle and twisting a nut on a bolt), the interplay of the force-based constraints and the geometric and discrete choices. %In particular we show how the system can consider strategies that are combinatorial assemblies of the actions and choose a satisfying strategy based on the parameters of the environment.  % considering removing the second sentence..
    \item Enable the planner to choose strategies that are robust by formulating this as cost-sensitive planning. We demonstrate how accounting for robustness impacts the robot's strategy.
\end{itemize}

% Could add this back when once order fixed, but id be happy to skip it
%After reviewing related work in \sref{sec:related_work}, we define our problem class and example domains in \sref{sec:approach}.
%\sref{sec:force} details the control strategy and mechanics models used within our planner.
%These methods are then contextualized within the example domains in \sref{sec:demos} where we also explore the types of solutions the planner finds.
%We take a step toward robust planning in \sref{sec:robust} by using cost-sensitive planning to find plans that are likely to succeed during open-loop execution.

% !TEX root = main.tex

\section{Background and Related Work}
\label{sec:related_work} 

Our goal of finding a discrete sequence of actions parameterized by continuous values lies at the heart of multi-modal motion planning (MMMP)~\cite{hauser2010multi,hauser2011randomized} and task and motion planning (TAMP)~\cite{garrett2020integrated}. 
MMMP plans motions that follow modes, e.g. moving through free space, and motions that switch between discrete modes, e.g. grasping, where each mode is a submanifold of configuration space.
%Solving an MMMP problem requires planning a valid mode sequence and planning paths within each mode.
TAMP extends MMMP by incorporating non-geometric state variables and a structured action representation that supports efficient search~\cite{garrett2017sample,kaelbling2011hierarchical,toussaint2015logic}.
Most, although not all~\cite{toussaint2020describing}, TAMP algorithms have focused on collision and kinematic constraints;
this paper focuses on integrating force-based reasoning with an existing TAMP method~\cite{garrett2020pddlstream}.

Most similar to our work, Toussaint \etal~formulates force-centric constraints that integrate into a path-optimization framework (LGP) for manipulation tasks~\cite{toussaint2020describing}.  
While LGP can search over strategies, in this paper the strategy was provided and fixed. 
While Toussaint \etal~take a more generic approach to representing interaction, their use of 3D point-of-attack (POA) to represent 6D wrenches prevents the system from considering patch contact, which is critical to our tasks. 
Levihn and Stilmann present a specific planner that reasons over which combination of objects in the environment will yield the appropriate mechanical advantage for unjamming a door~\cite{levihn2014using}. 
The type of the door directly specifies which strategy to use (lever or battering ram) and the planner considers the interdependencies of force-based and geometric-based decisions for each application.
%Kresse and Beetz link symbolic reasoning with force-control primitives, although not in the context of forceful manipulation~\cite{kresse2012movement}.  

Michelman and Allen formalize opening a childproof bottle via a finite state machine, where the overall strategy and some of the continuous parameters, such as the grasp, are fixed~\cite{michelman1994forming}. 
Holladay \etal~consider force- and motion-based constraints in planning a fixed sequence of actions to enable tool use~\cite{holladay2019force}.  
While these systems reason over geometric and force constraints, these constraints do not impact the sequence of actions, i.e. the choice of strategy. 

Several recent works have characterized types of force-based motions. 
Gao \etal~defined ``force-relevant skills'' as a desired position and velocity in task space, along with an interaction wrench and task constraint~\cite{gao2019learning}. 
Mahschitz \etal~termed ``sequential forceful interaction tasks'' as those characterized by point-to-point movements and an interaction where the robot must actively apply a wrench~\cite{manschitz2020learning}.  Chen \etal~define ``forceful operations'' as a 6D wrench $f$ applied at a pose $p$ with respect to a target object~\cite{chen2019manipulation}. 
In this paper, we adopt their definition of forceful operations to characterize the type of interactions our system plans for. 
Chen \etal~focus on finding environmental and robot contacts to stabilize an object while a human applies the forceful operation. 
In our work, a robot must both stabilize the object and apply the forceful operation. 

Stabilizing, or fixturing, an object is a common requirement of any system forcefully operating on those objects. 
The goal of fixturing is to fully constrain an object or part, while enabling it to be  accessible~\cite{asada1985kinematics}.
Fixture planning often relies on a combination of geometric, force and friction analyses~\cite{hong1991fixture}.
There are various methods of fixturing including using clamps~\cite{mitsioni2019data}, using another robot to directly grasp or grasp via tongs~\cite{watanabe2013cooking,stuckler2016mobile,zhang2019leveraging}, or using the environment and relying on friction, such as in fixtureless fixturing or shared grasping~\cite{chavan2018regrasping,hou2020manipulation}. 
Within this paper we consider fixturing via grasping and environmental contacts. 

% !TEX root = main.tex

\section{Forceful Manipulation}  % problem

In this section, we characterize the class of problems considered in this paper and we introduce two domains that illustrate the class.

\begin{figure*}[t!]
  \begin{subfigure}[t]{0.24\textwidth}
    \includegraphics[width=\textwidth]{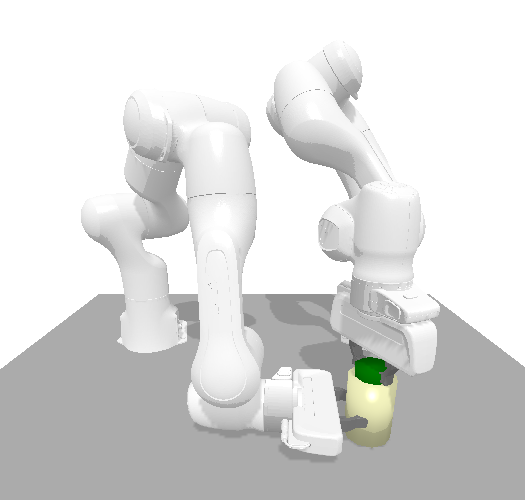}
    \caption{GT + RF} \label{fig:lidgrid_grasp_arm}
  \end{subfigure}
  \begin{subfigure}[t]{0.24\textwidth}
    \includegraphics[width=\textwidth]{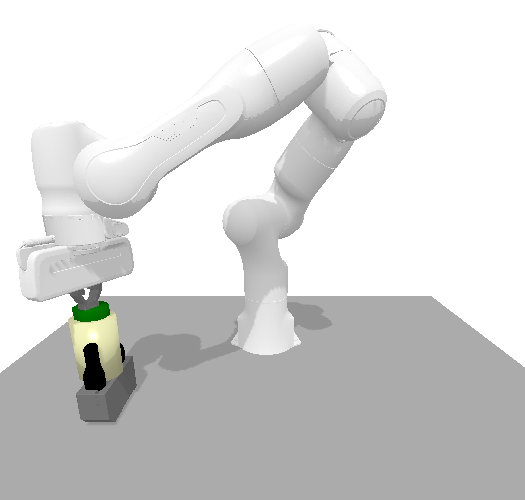}
    \caption{FT + VF} \label{fig:lidgrid_finger_vise}
  \end{subfigure}
  \begin{subfigure}[t]{0.24\textwidth}
    \includegraphics[width=\textwidth]{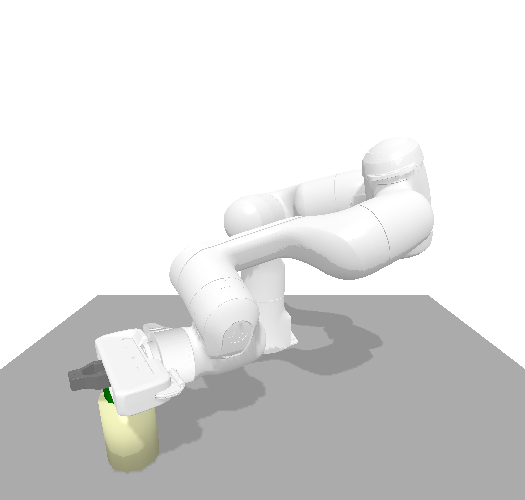}
    \caption{PT + SF(T)} \label{fig:lidgrid_palm_surface}
  \end{subfigure}
  \begin{subfigure}[t]{0.24\textwidth}
    \includegraphics[width=\textwidth]{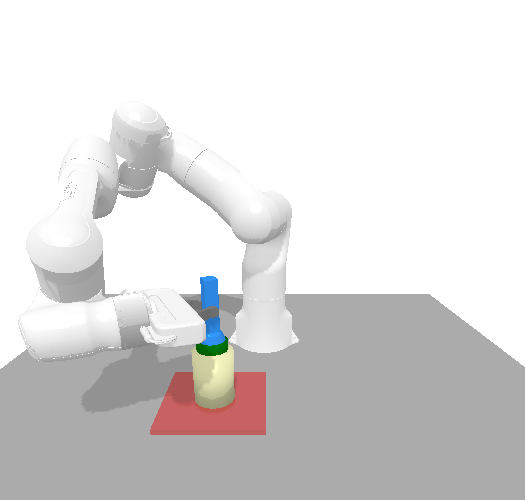}
    \caption{TT + SF(M)} \label{fig:lidgrid_tool_surface}
  \end{subfigure}
\caption{To open a childproof bottle, the robot must forcefully push and twist the lid, while fixturing the bottle. We define four strategies for twisting the lid: grasping on the lid (GT), pushing through the fingertips (FT) or the palm (PT) or grasping a pusher-tool that contacts the lid (TT). The system may fixture the bottle via another robot's grasp (RF), a vise's grasp (VF) or by securing it between the table (SF(T)) or a high-friction rubber mat, shown in red, (SF(M)).} \label{fig:lidgrid}
\end{figure*}

\subsection{Problem Definition}
\label{sec:problemDef}

We are interested in solving forceful manipulation tasks via a sequence of high-level actions, i.e. a strategy.
Each action is implemented by a parameterized controller and is associated with constraints relating the continuous parameter values, such as robot configurations, poses, paths, etc. that must be satisfied for the controller to achieve its desired effect.  The choice of a strategy requires finding satisfying values for all of these constraints.

Borrowing from Chen et al., we are interested in tasks that  involve \textit{forceful operations}, where the robot exerts a 6D wrench ($[f_{x}, f_{y}, f_{z}, t_{x}, t_{y}, t_{z}]$) on an object~\cite{chen2019manipulation}. 
%We consider quasi-static tasks, where the forces are always in balance and produce relatively slow motions.  (was going to add, but its in the assumptions below)
Actions that perform forceful operations (1) require the robot to use a controller that exerts force and (2) plan for the robotic system to be able to exert that force. 
We view the robotic system, composed of the robot joints, grasps and other possible frictional contacts, as a \textit{forceful kinematic chain}. 
To exert a wrench, the forceful kinematic chain must be maintained, i.e. each joint must be stable under the imparted wrench. 
We informally use the word stable to refer to stable equilibrium of the forces and torques at all the ``joints''.
%must be strong enough to withstand the imparted wrench. 
Likewise, if we are exerting force upon an object, we need to fixture that object, hence creating a second forceful kinematic chain. 
In \sref{sec:approach} we explain how to incorporate stability into the planner as a constraint and then describe several types of joints in forceful kinematic chains and their mathematical models.

In this paper we assume a quasi-static physics model and, as input, are given geometric models of the robot, the objects and the environment along with the poses of each object.      
Physical parameters, such as the object's mass and center of mass, and friction coefficients, are known.
%However, the ability to exert wrenches is not sufficient for success. %; the system's kinematic chain must satisfy various force-based constraints. 
%The robotic system that is exerting this force is made up of a series of links, creating a kinematic chain. 
%In the simplest case, these links are the joints of the robot. 
%However, as seen in \figref{fig:fig1}(top left), if the robot is exerting a force through a grasped tool, that grasp is an additional link within the chain. 
%In order to successfully exert a wrench, the kinematic chain must be maintained, i.e. each link must be strong enough to withstand the imparted wrench. 
%Likewise, if we are exerting force upon an object, we need to stabilize or fixture that object. 
%One way to consider this is that we need a second kinematic chain, with its own set of links that anchors the object.

%Both cases involve, for a given possible kinematic chain, verifying whether each link in the chain is strong enough to withstand the exerted force. 

\subsection{Example Domains}
\label{sec:domains}

To ground our work in concrete problems, we consider two example domains: (1) opening a childproof bottle and (2) twisting a nut on a bolt (\figref{fig:lidgrid} and \figref{fig:nutgrid} respectively). 
For each domain, we define the forceful operation that represents the task, what object(s) must be fixtured, and a set of possible actions for imparting that forceful operation and fixturing the objects. 
In both domains we also include generic actions such as: move, move while holding, pick, and place. 

%In both domain we include generic actions such as \texttt{move} (representing a joint-space path from one configuration to another), \texttt{move\_holding} (the move action while holding an object), \texttt{pick} (grabbing an object from a surface) and \texttt{place} (putting an object on a surface).
%We briefly overview the domain-specific actions below, which are further detailed, along with their force-based constraints, in \sref{sec:demos}.

\subsubsection{Childproof Bottle Opening}

In the first domain, the objective is to open a push-and-twist childproof bottle, as introduced in \sref{sec:intro}.
%We specify the goal as removing the lid, which accomplished via the action \texttt{remove-cap}.
We specify the push-twisting, required before removing the lid, as the forceful operation of applying wrench $(0, 0, -f_{z}, 0, 0, t_{z})$ in the frame of the lid (\figref{fig:fig1}-bottom left), where we assume $f_{z}$ and $t_{z}$ are given.
While performing this action, the robot must fixture the bottle to prevent its motion. 
We define four push-twisting actions and three fixturing methods.

%\sout{All of the forceful operation actions involve a guarded move to establish contact~\cite{will1975experimental}, followed by a controller that exerts force, as described in \sref{sec:controller}.}
Each of the four push-twisting actions, illustrated in \figref{fig:lidgrid}, vary in how the robot contacts the lid and thus the construction of the forceful kinematic chain: grasping on the lid, pushing through fingertips or the palm or grasping a pusher-tool that contacts the lid.
We refer to these push-twisting strategies as GT (grasp-twist), FT (finger-twist), PT (palm-twist) and TT (tool-twist). 
For the last three actions, the robot can reason over applying additional downward force, since this can safely increase the stability of the action. 
%For the last three operations, the robot can reason over applying more downward force, in addition to what is required by the task. 

%Each of these contacts are circular with an even pressure distribution and thus their stability is modeled by the limit surface from \sref{sec:limit_surface}, but with varying parameters.
%However the parameters, the friction coefficient $\mu$, radius $r$ and normal force $N$ vary for each contact. 
%The first is \texttt{grasp-twist}, where the robot directly grasps the cap with its parallel jaw gripper, where the normal force $N$ is the commanded gripping force. 
%We also consider three variants where the robot arm makes contact with the cap without explicitly grasping it: contact via the fingertips, the palm and a pusher-tool (\texttt{finger-pushtwist}, \texttt{palm-pushtwist}, \texttt{tool-pushtwist}), as seen in \figref{fig:lidgrid}.
%For these actions the robot can reason over applying more downward force (in addition to what is required by the task), since, given the ellipsoidal model in \eref{eqn:limit_surface}, increasing the applied normal force $N$ by applying more force enlarges the space of wrenches that can be resisted.  

%The \texttt{tool-pushtwist} action has the additional constraint that the grasp on the tool must also be stable, which also modeled with the limit surface.
%This can present a tradeoff, where applying more downward force increases the stability of cap-tool contact while increasing the wrench that the grasp tool-robot contact must sustain. 

There are three methods to fixture the bottle, as shown in \figref{fig:lidgrid}: grasping it with another robot, grasping it with a vise (here implemented via a rigidly mounted robot hand) or exerting additional downward force to secure the bottle with friction.  
For the last method, the frictional surface can either be the table, or a high-friction rubber mat. 
We refer to these fixturing strategies as RF (robot-fixture), VF (vise-fixture), SF(T) (surface-fixture using the table) and SF(M) (surface-fixture using the mat). 

%Either a second robot or a vise (here implemented via a rigidly mounted robot hand) can grasp the bottle (\texttt{robot-fixture} and \texttt{vise-fixture} respectively), where this grasp must be stable according to the limit surface. 
%In the third fixturing strategy, like with the pushtwists, we can apply additional downward force such that, between the contact and the table, the bottle is immobilized. 
%Since the bottom of the bottle is circular, we again model this stability with the limit surface where the normal force $N$ is the combination of downward planned applied wrench, the additional downward force and force due to gravity. 

\begin{wrapfigure}{r}{0.35\columnwidth}
\vspace{-5.5mm}
\centering
   \includegraphics[width=3cm]{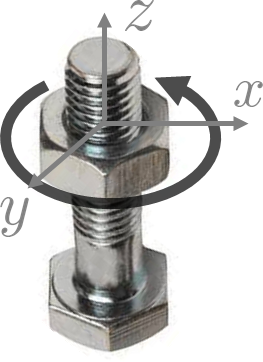}
   \caption{In the nut twisting domain, the forceful operation is to apply a torque in z, in the frame of the nut.}
  \vspace{-1mm}
\label{fig:nut_task_wrench}
\end{wrapfigure}

\subsubsection{Nut Twisting}

In the second domain, the robot twists a nut on a bolt by applying the wrench $(0, 0, 0, 0, 0, t_{z})$ in the frame of the nut (\figref{fig:nut_task_wrench}), with $t_{z}$ given as input, while fixturing the beam holding the bolt.
%At this point, we do not consider the harder, more general task of twisting the nut until it is tight, although this 
We do not consider the more general task of twisting a nut \textit{until} it is tight.
This version of the task, which would require monitoring and re-planning, is discussed as future work in \sref{sec:discussion}

\begin{figure}[t!]
\centering
  \begin{subfigure}[t]{0.49\columnwidth}
    \includegraphics[width=\textwidth]{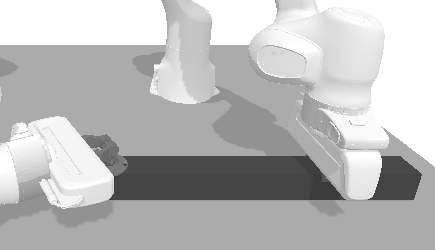}
    %\caption{\texttt{grasp-twist} with \\ \texttt{robot-fixture}} \label{fig:nutgrid_hand_arm}
  \end{subfigure}
  \begin{subfigure}[t]{0.49\columnwidth}
    \includegraphics[width=\textwidth]{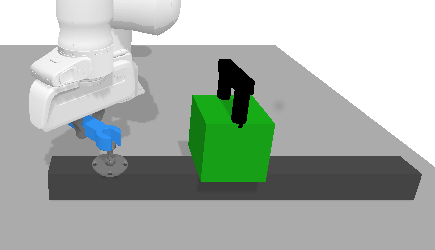}
    %\caption{\texttt{spanner-twist} with \\ \texttt{weighted-fixture}} \label{fig:nutgrid_tool_weight}
  \end{subfigure}
\caption{To twist a nut on the bolt, the robot can use either its fingers or a spanner (in blue). While twisting, the robot must fixture the beam that the bolt is attached to. The robot can fixture either via another robot's grasp or by weighing down the beam with a large mass (in green).} 
\label{fig:nutgrid}
\end{figure}

The robot can either twist the nut, executing a controller that exerts force, with its fingers or by using a spanner (\figref{fig:nutgrid}). 
The beam can be fixtured either by having a second robot securely grasp it or by placing a heavy block on the beam to weigh it down.
\newline
%The robot can either twist the nut by directly grasping it (\texttt{grasp-twist}) or by using a spanner, where the robot grasps the spanner and the spanner ``grasps'' the nut (\texttt{spanner-twist}).
%The stability of each of these grasps are verified by the limit surface, although for the spanner's grasp on the nut the normal force $N$ is zero. 
%Therefore this grasp is only stable if all non-negative elements of the wrench exist in the three dimensions that are stabilized by the geometry of the spanner. 

%One fixturing strategy is to have another robot grasp the bar (\texttt{robot-fixture}), with its grasp stability checked via the limit surface. 
%The second fixturing strategy is to place a weight on the bar in order to weigh it down (\texttt{weighted-fixture}). 
%We model the contact between the bar and the table with the generalized friction cone, placing friction cones at each of the four corners of the rectangular contact, as seen in \figref{fig:frictionCone}.
%The generators of the friction cones are scaled by the normal force $N$, which is determined by the mass and location of the weight.
%We model the weight's normal force distribution as simply supported 1D beam with partially distributed uniform load (\rhnote{dont know what to cite for this. Have math in \sref{sec:beam}}). 

%In each domain the variety of twisting and fixturing methods gives rise to a large set of possible strategies to search over. 
%We explore how the various geometric and force constraints impact this search in \sref{sec:demos}.
%First, we outline the underlying mechanics that capture the force-based constraints.
%\newline
In each domain, the variety of push-twisting (or twisting) and fixturing methods gives rise to a combinatorial set of possible strategies to search over. 
For example, one such strategy for the childproof bottle domain that uses the grasp-twist (GT) and robot-fixture (RF) methods, as partly visualized in \figref{fig:lidgrid_grasp_arm}, is: 
\texttt{move(robot$_{0}$, path$_{0}$)}, \texttt{pick(robot$_{0}$, bottle, grasp$_{0}$)}, \texttt{move(robot$_{1}$, path$_{1}$)}, \texttt{grasp-twist(robot$_{1}$, bottle, path$_{2}$, wrench)}, \texttt{move(robot$_{1}$, path$_{3}$)}, \texttt{pick(robot$_{1}$, cap, grasp$_{0}$)}, where some of the parameters of the actions have been omitted for brevity. 
The goal of the planner, described in \sref{sec:approach}, is to both search through the set of feasible strategies (action sequences) and solve for the parameters of each action, all while respecting the geometric and force-based constraints. 

%We provide simulation videos showing all combinations of twist strategies and fixture strategies in both  domains at:  \href{https://mcube.mit.edu/forceful-manipulation/}{https://mcube.mit.edu/forceful-manipulation/}. We also show real-robot executions of some of the combinations.

% !TEX root = main.tex

\section{Force-Based TAMP}
\label{sec:approach}

We next detail in \sref{sec:planning} how we introduce force-based reasoning into an existing TAMP method, including adding a new type of variable, new controllers and new constraint functions, and how we extend it to generate robust strategies via cost-sensitive planning.
We then discuss in \sref{sec:force} the controller for exerting wrenches and the mathematical models used to verify forceful kinematic chains.

\subsection{Planning Approach}
\label{sec:planning}

To solve for both the strategy and the parameters of the strategy in a forceful manipulation problem, we use PDDLStream, a task and motion planning algorithm~\cite{garrett2020pddlstream} that has been used in a variety of robotics domains, including pick-and-place in observed and partially-observed setting~\cite{LIS267}.

\subsubsection{Extending PDDLStream}

PDDLStream solves hybrid (discrete/continuous) problems by sampling, in a focused fashion, the continuous parameters and generating and solving a sequence of discrete planning problems until a solution to the original hybrid problem is found.  
The focused sampling is achieved by combining a small set of samplers that are conditional on the output of other samplers, such as a sampler for kinematic solutions conditioned on sampled poses of objects.  

We define each of our example domains using the PDDLStream planning language, an extension of the PDDL language~\cite{mcdermott1998pddl}, by specifying variable types (e.g. poses), predicates (e.g. HandEmpty) and actions (e.g. pick) that are relevant to the domain.
Each action is a parameterized controller and is characterized by preconditions on the state, or requirements for executing that controller, and its resulting effects to the state. 
Given this specification, along with an initial state and goal state, PDDLStream finds a sequence of actions and their parameters. %, whose parameterize controllers are then executed by the robot. 

In order to leverage PDDLStream for forceful manipulation we add a wrench type, define actions that exert wrenches and introduce force-based constraints.  
First, we add \texttt{(Wrench ?w)} as a state variable type, where \texttt{?w} is a 6D wrench, specified with respect to a reference frame, which is usually attached to an object.
This addition enables wrenches to be directly and easily incorporated into the specification.
%For example, we can now specify actions whose effect is exerting a wrench or add the precondition that any grasp selected by be able to resist a certain wrench application.

Actions and their controllers which exert the forceful operations, such as the push-twist actions in the childproof lid domain, are parameterized, in part, by this wrench variable and we can characterize an effect of the action as accomplishing the forceful operation. 
Additionally, a precondition for executing these wrench-exerting actions, as mentioned in \sref{sec:problemDef}, is that the forceful kinematic chain is maintained and that the objects the robot is acting upon are fixtured.
This constraint propagates the planned exerted wrench through the joints of the relevant forceful kinematic chain and is deemed satisifed if all the joints (robot and contacts) are stable with respect to the application of that propagated wrench. % had: this constraint, accomplished via the set of samplers, propagates the planned 

For example, when generating a grasp, the grasp sampler transforms the exerted wrench into the grasp frame and only considers the grasp feasible if it is stable.
\sref{sec:controller} discusses the controller that exerts wrenches and \sref{sec:kinematicChain} details the forceful kinematic chain constraint. 
%Verification of this constraint is accomplished via samplers, discussed in \sref{sec:kinematicChain}. 

%The constraints first propagate the planned exerted wrench through the joints of the relevant forceful kinematic chain. 
%The constraint is satisfied if the joint is stable with respect to the application of that propagated wrench.
%In this context, a contact is deemed stable if the application of a wrench does not produce resultant motion, i.e. the contact "sticks"~\cite{}.
%For links in the chain that involve frictional contact, like grasps, we deem them stable if the contact can transmit the wrench without slipping.
%For example, consider the two forceful kinematic chains within \figref{fig:nutgrid} (right).
%For example, when generating a grasp, the grasp sampler propagates the exerted wrench into the grasp frame and only considers the grasp feasible if it is stable. 
%To assess whether the block weighs down the bolt structure enough to properly fixture it, we add a test sampler, which is a sampler that certifies whether a constraint is met. 

%\tlpnote{Describe how the kinematic chain constraints show up in the system.  Were there any new samplers needed?}

%The parameterized controller for these actions that exerts the wrench is detailed in \sref{sec:controller}.

\subsubsection{Robust Planning}
\label{sec:robust}

Given the ability to generate plans that satisfy geometric- and force-based constraints, we now aim to produce \textit{robust} plans.
We search for plans that maximize the probability of succeeding during open-loop execution. 
We focus on protecting against stability-based failures along the forceful kinematic chains due to uncertainty in physical parameters. 
For example, we want to discourage the system from selecting a grasp where a small change in the friction coefficient would break the stability of the grasp. 

We formulate this as cost-sensitive planning, where the cost of an action is given as $-\log$(Pr[success(action)]). 
Minimizing this cost is equivalent to maximizing the plan success likelihood. 
We define the probability of action success by sampling 100 sets of parameter values, each with a random epsilon perturbation, and evaluating our models (detailed in \sref{sec:kinematicChain}) to assess the stability of the forceful kinematic chain. 
We perturb, when applicable, parameters such as the friction coefficient, the planned applied wrench, the contact frame, the effective size of the contact patches, etc. 
By reasoning over perturbations in the parameters of the problem, we assess the robustness of the forceful kinematic chain.

\subsection{Constraints on Forceful Operations}
\label{sec:force}

We next discuss the controller used to exert wrenches and the mechanics models used to asses the stability of forceful kinematic chains.
%the modeling tools from mechanics used to reason over the system's stability.
While we overview a few types of joints, and their corresponding models, alternative joints or models could easily be integrated. 

\subsubsection{Exerting Force}
\label{sec:controller}

Forceful interaction involves exerting a wrench at a pose. 
While there are several control methods for this explicitly, such as force control~\cite{zeng1997overview} or hybrid position-force control~\cite{mason1981compliance,hou2019robust}, we opt to use a Cartesian impedance controller.
Cartesian impedance control treats the interplay of interaction forces and motion deviations as a mass-spring-damper system~\cite{hogan1985impedance}. 
We can exert forces and torques by offsetting the target Cartesian pose and adjusting the impedance parameters~\cite{kresse2012movement}.
We chose to only vary the stiffness parameter, $K_{p}$, and set the damping parameter, $K_{d}$, to be critically damped, i.e. $K_{d} = 2\sqrt{K_{p}}$.

%We experimentally characterized the force exerted by the manipulator in a direction given the pose offset and stiffness in that direction. 
To have a degree of open-loop control over the magnitude of the exerted wrench, we characterized experimentally the relation between wrench magnitude, pose offset and stiffness.  
During planning, we use this experimental relation to select, given the stiffest possible setting, the desired pose offset. 

%\sout{While impedance control sacrifices precision with respect to force tracking, precise force control is not required for the tasks we are considering.  Additionally, impedance control's compliant behavior is safer in our open-loop setting.}

\subsubsection{Maintaining Forceful Kinematic Chain}
\label{sec:kinematicChain}

With the actions that exert force, the planner must ensure that each joint in the forceful kinematic chain is stable.
%In considering the forceful operation, the planned exerted wrench is propagated through each link in the robot's kinematic chain and the fixture's kinematic chain.
%Along each propagation, we assess whether the link is stable with respect to the application of that wrench. 
For our two domains we consider several classes of joints for which we model the space of wrenches they can withstand. 

%The first two classes of links are planar contact between two rigid bodies. 
%Given a line or patch contact, we can model the frictional wrench provided by the contact via the generalized friction cone~\cite{erdmann1994representation}. 
%In the case of a circular patch contact with a uniform pressure distribution, we model the frictional wrench via the limit surface~\cite{goyal1991planar}. 
%In both cases we are considering planar contact, where the frictional wrenches are three dimensional. 
%%Therefore after transforming our 6D wrench to the appropriate frame, we extract the appropriate 3D slice for either models. 
%We will assume that any wrenches in the other three dimensions are resisted kinematically.
%For example, consider the robot's grasp on the tool in \figref{fig:lidgrid}(far right). 
%In three dimensions the motion of the tool in-hand is prevented by the geometry of the hand. 
%The mathematical models, either the generalized friction cone or limit surface, describe whether the friction resulting from the grasp prevents motion in the other three directions. 
%The last class of links are the robot's joints, where the bounds on these joints are expressed via the joint torque limits. 
%%Here the full 6D wrench is considered. 

\begin{figure}[t]
\centering
    \includegraphics[width=0.95\columnwidth]{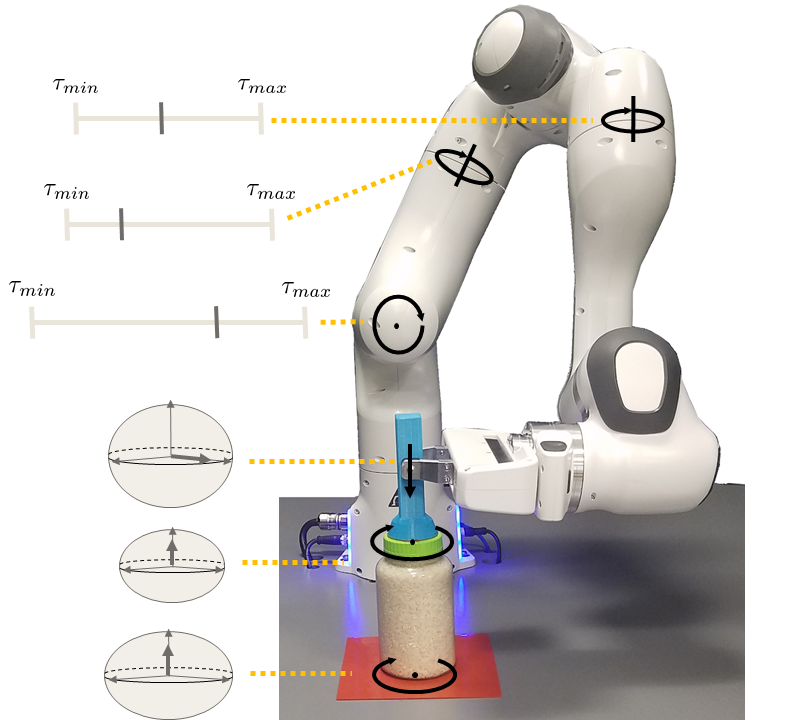}
\caption{Along each joint of the forceful kinematic chain, we first project the expected wrench (in black) into the subspace defined by each joint and then verify if the joint is stable under that wrench. Each type of joint connects to a mathematical model that visualizes this space. For circular patch contacts, we check the friction force against a limit surface ellipsoidal model and for each robot joint we check against the 1D torque limits. }
\label{fig:wrenchSpaces}
\end{figure}

For frictional planar joints, the friction wrenches are 3D, and we represent the boundary of the set of frictional wrenches in the three dimensional friction subspace of the plane of contact with a limit surface~\cite{goyal1991planar}.
We use two ways to approximate the limit surface, depending on the characteristics of the planar joint (see a) and b) below). 
In the other three dimensions, we assume any wrenches are resisted kinematically by non-pentration reaction forces. 
For example, in \figref{fig:wrenchSpaces}, the joint between the gripper and tool is maintained along three axes by friction and in the other three axes by the gripping force.
%Of the three axes maintained by friction in the example in the figure, there is only one (vertical downward force) that will see significant force.
%in-hand motion of the tool in-hand in prevented in three dimensions by the geometry of the robot's grasp. 
For the robot's joints, the set of wrenches that can be transmitted are directly bound via the joint torque limits. 

\paragraph{Limit Surface for Small Circular Patch Contacts}
For small circular patch contacts with uniform pressure distributions (e.g. the fingers-bottle, bottle-surface contacts), we leverage the ellipsoidal approximation of the limit surface~\cite{xydas1999modeling}.
The ellipsoid is centered in the contact frame, $w = [f_{x}, f_{z}, m_{y}]$, and, for isotropic friction, is defined by $w^{T}Aw~=~1$ where:
\[A~=~ 
\begin{bmatrix}
    \frac{1}{(N\mu)^2} &                    & 0  \\
                       & \frac{1}{(N\mu)^2} &   \\
    0                  &                    & \frac{1}{(Nk\mu)^2} 
\end{bmatrix}
\]
%$A= Diag \big( \frac{1}{(N\mu)^2}, \frac{1}{(N\mu)^2}, \frac{1}{(Nk\mu)^2} \big)$ 
such that $\mu$ is the friction coefficient, $N$ is the normal force and, given our assumptions, $k \approx 0.6r$ where $r$ is the radius of the contact~\cite{xydas1999modeling,shi2017dynamic}.
Having propagated the exerted wrench into the contact frame, we check if this wrench lies within the ellipsoid, which would indicate a stable contact:
\begin{equation}
\frac{f_{x}^{2}}{(N\mu)^{2}} + \frac{f_{z}^{2}}{(N\mu)^{2}} + \frac{m_{y}^{2}}{(Nk\mu)^{2}} < 1
\label{eqn:limit_surface}
\end{equation}

%Therefore if \eref{eqn:limit_surface} is true, the contact is stable. 
%Given the geometry of the robot's hands, we use the limit surface to model the stability of all of the grasps and the contact made via the robot's palm and fingertips. 
%In the childproof bottle domain, because the bottom of the bottle is circular, we also model the contact between the bottle and the surfaces with the limit surface. 

%\paragraph{Generalized Friction Cone}
\paragraph{Limit Surface for More General Patch Contacts}
For contacts with more general shapes and less uniform pressure distributions, we directly model the contact patch as a set of point contact, each with its own normal force (localized pressure) and its own friction limits. 
%we directly construct a polyhedral approximation of the limit surface.
%We compute an approximation to the pressure distribution by distributing it among the corners of the contact patch, and then compute the convex hull of the generalized friction cones at each corner. 
Given a contact patch we model the force it can transmit as the convex hull of generalized friction cones placed at the corners of the patch.
Generalized friction cones, based on the Coulomb friction model, represent the frictional wrench that a point contact can offer~\cite{erdmann1994representation}. 
%This generates the limit surface. And then explain what generalized friction cones are, and how to combine them, as we currently have in a)
%Erdmann introduced the generalized friction cone, based on the Coulomb friction model, for representing frictional wrench that point contact can offer~\cite{erdmann1994representation}. 
%To represent the frictional force afforded by a patch (line) contact, we place multiple point contacts over the patch (along the line) and take the convex hull of the generalized friction cone at each of the point contacts. 
%In our context, if the planned exerted wrench, in the reference frame of the patch contact, lies within this convex hull, the frictional force can resist the wrench and the contact is stable. 
We represent the friction cone, FC, at each point contact with a polyhedral approximation of generators:
\begin{equation} 
FC = \{ (\mu, 0, 1), (-\mu, 0, 1), (0, \mu, 1), (0, -\mu, 1) \}
\end{equation}
for a friction coefficient, $\mu$~\cite{lynch2017modern}. 
These generators can be scaled by the applied normal force. 
%\figref{fig:frictionCone} visualizes the four friction cones representing the patch contact between the bar and the table. 
Given this approximation, the generalized friction cone can be written as: 
\begin{equation}
V = \{ v = J_{f}^{T} F~|~F \in {FC} \}
\label{eqn:generalized_friction_cone}
\end{equation}  
where $J_{f}^{T}$ is the Jacobian that maps contact forces $f$ from the contact frame, where FC is defined, to the object frame. 
If the exerted wrench, in the reference frame of the patch contact, lies within the convex hull of $V$, the frictional wrench can resist the exerted wrench and the contact is stable.

In the nut-twisting domain (\figref{fig:nutgrid}), we use the generalized friction cone to model the contact between the table and the beam holding the bolt, placing friction cones at the four corners of the beam.  
%the contact between the bolt structure and the table in assessing the stability of fixuring via heavy weight. 
%We represent this patch contact by placing friction cones at each of the four corners.
In evaluating the stability of fixturing via a heavy weight, the applied normal force, determined by the mass and location of the weight, is modeled as a simply supported 1D beam with a partially distributed uniform load.

\paragraph{Torque Limits} 
The last type of joint we consider are the joints of the robot, where the limit of each joint is expressed via its torque limits. 
We relate the wrenches at the end effector to robot joint torques through the manipulator Jacobian, $J_{m}$.
Specifically, given a joint configuration $q$ and wrench $w$, the torque $\tau$ experienced at the joints is modeled by $\tau = J_{m}^{T}(q)w$.
The forceful kinematic chain is stable if the expected vector of torques $\tau$ does not exceed the robot's torque limits $\tau_{lim}$: 
\begin{equation}
J_{m}^{T}(q)w_{ext} < \tau_{lim}.
\label{eqn:torque_limit}
\end{equation}
\newline
In summary, we have three models, each of which captures the spaces of wrenches that a particular joint can resist. 
The frictional contact models create ellipsoids and polytopes in wrench space. 
The torque limit creates a $n$-d box in torque space, where $n$ is the number of joints, which can be thought of as a series of 1D bounds for each joint. 
\figref{fig:wrenchSpaces} visualizes how these varying models are used to asses the stability of the forceful kinematic chains.  

% !TEX root = main.tex
\section{Empirical Evaluation}
\label{sec:demos}

\begin{figure*}[th!]
\centering
    \includegraphics[width=0.33\textwidth]{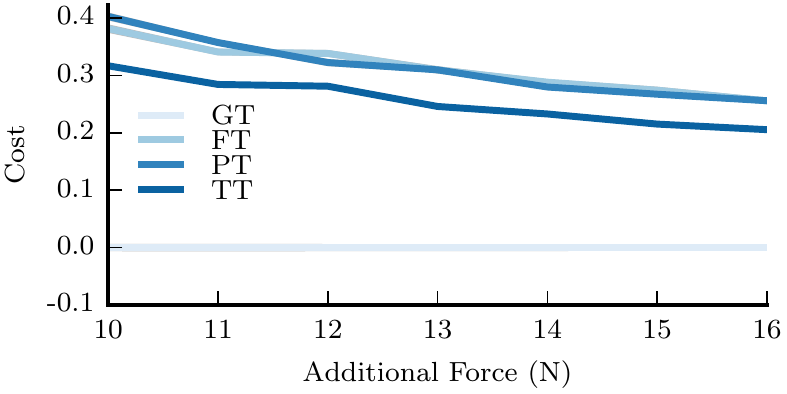}
    \includegraphics[width=0.33\textwidth]{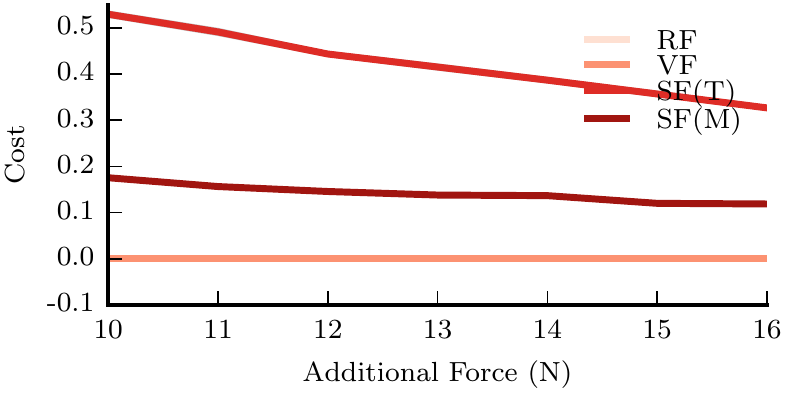}
    \includegraphics[width=0.33\textwidth]{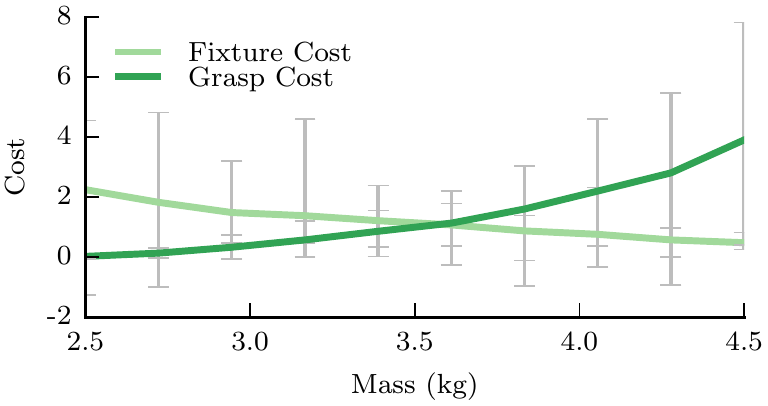}
\caption{Robustness Evaluation. In the childproof bottle domain, we evaluate the cost of each twisting method (left) and fixture method (center) across varying values of downward force. For each point, we average one hundred samples, randomizing over the grasp (when applicable). The standard error is plotted, but minuscule. From our cost definition, a higher cost corresponds to a less robust action. In the nut-twisting domain (right) we consider the trade-off in the grasp cost versus the fixturing cost. At each weigh value, we randomly sample the pose of the weight along the beam and the grasp on the weight. Since, at the extremes, some costs evaluate to infinity, we plot the median and a 95\% confidence interval.}
\label{fig:robustness}
\end{figure*}

We provide a few illustrative examples of how the system reasons over strategies.
We also further demonstrate the system in the supplemental video~\cite{ftampWebpage}.
%, adjusting to variations in environmental parameters. 
%When searching over strategies, PDDLStream begins with the shortest strategies, in terms of action length, and lengths its search horizon broadens to longer strategies if it can't find a satisfying solution. 
\subsection{Breadth of Solutions}
We conduct two ablation studies in the childproof bottle domain (\tref{table:ablation}). 
%During each run, the bottle's location is randomized. 
In, the first study, A1, we search over all possible fixturing strategies, with the twisting strategy fixed (GT, grasp-twist). 
In A1, the bottle starts at a random location on the table. 
%\rhtodo{Considering that the planner first for the strategy with the fewest number of actions, we incrementally remove these shorter strategies to demonstrate the longer ones.} 
%\rhtodo{For example, to show GT+SF(M), we removed the second robot arm such that GT+RF was not a feasible strategy. }
In the second study, A2, we search over all possible twisting strategies, starting the bottle on the rubber mat to use the surface fixturing strategy (SF(M), surface-fixturing on the mat). 
%Like in the fixturing case, we remove the shorter strategies to explore the longer strategy (TT, tool-twist). 

%Because the planner biases towards plans with the fewest number of actions, 
Because the underlying search over strategies within PDDLStream biases towards plans with the fewest actions, we incrementally invalidated the shorter strategies in order to force exploration of the alternative, longer strategies. 
For example, in A1 we invalidated fixturing via the table or a robot grasp (GT+SF(T) and GT+RF) as feasible strategies by decreasing the friction coefficient between the bottle and the table and by removing the second arm, respectively. 
Accounting for this, the planner chose the longer, feasible strategies of fixturing with either the vise (GT+VF) or with the rubber mat (GT+SF(M)).

%('R ', 59.738388204580005, 9.408764223815794, ' vf ', 95.265781354840001, 35.044098489196251, ' sf ', 141.89954099664001, 72.903228523282451, 'sf_table ', 177.17463579073998, 51.160716232152225)

%('GT ', 36.821242856979993, 1.7766032142478232, 'PT', 25.239777803399999, 3.118236286517055, 'FT', 63.362480211200001, 34.974588332781757, 'TT ', 40.251920175560002, 5.1329978045245364)

\begin{table}[t!]
 \centering
 \begin{tabular}{l c c}%{c  c  c | c} 
 \hline 
 \multicolumn{3}{c}{Ablation \# 1} \\ [0.5ex] 
 \hline\hline
 Strategy & \# Steps & Planning Time  (SE) \\% & Comments \\ 
 \hline
 GT+SF(T)* & 4 & 177 (51) \\ %& if high $\mu$  \\ 
 \hline
 GT+RF & 6 & 60 (9.4) \\ %&  \\ 
 \hline
 GT+SF(M)** & 8 & 142 (73) \\ %& \multirow{2}{*}{if no shorter strategies} \\
 \cline{1-3}
 GT+VF** & 9 & 95 (35) \\ %&  \\ 
 \hline\hline 
 \multicolumn{3}{c}{Ablation \# 2} \\ [0.5ex] 
 \hline
 Strategy & \# Steps & Planning Time (SE) \\ %& Comments \\ 
 \hline\hline
 GT+SF(M) & 4 & 37 (1.8) \\ %&  \\ 
 \hline
 PT+SF(M) & 4 & 25 (3.1) \\ %& \\
 \hline %\cline{1-3}
 FT+SF(M)* & 4 & 63 (35) \\ %& if high $\mu$  \\
 \hline
 TT+SF(M)** & 8 & 40 (5.1) \\ %& if no shorter strategies \\
 \hline
\end{tabular}
\caption{For each ablation study, we provide the number of steps for each strategy and the average planning time in seconds (and standard error) over five runs. *: Utilized a higher friction coefficient $\mu$ to increase feasibility **: Invalidated shorter strategies to force to planner to find these longer strategies.}
\label{table:ablation}
\end{table}

\subsection{Robustness of Solutions}
We next consider how accounting for robustness impacts the choice of strategy. 
In \figref{fig:robustness}-left we evaluate each of the push-twisting methods in the childproof bottle domain, across varying magnitudes of additional downward force, a parameter the planner must choose. 
Except for the grasp twist (GT), all of the methods are fairly susceptible to perturbations in their parameters.  
%The grasp twist, who's stability is not impacted by this force, is shown to be very robust, in contrast to the other twist methods, 
Each of these methods decreases in cost, and therefore increases in robustness, with additional downward force. % as this expands their limit surfaces.
We repeat this analysis for the fixturing methods in the same domain (\figref{fig:robustness}-center), where the robot and vise fixturing methods (RF, VF) both have a cost of zero. 
In comparing fixturing with the table (SF(T)) and the mat (SF(M)), the mat's higher friction coefficient makes it more robust.
The robust planner is incentivized to avoid shorter but more brittle plans, such as fixturing with the table, and to opt for longer, more reliable plans. 
%In planning robustly biases us towards the twisting and fixturing method that are the least sensitive to perturbations in parameters and thus the most likely to succeed. 

In the nut-twisting domain we explore robustness by considering a scenario where the robot must chose between several weights, of varying mass, to fixture the beam with. 
For a given mass, we sample 100 placement locations along the beam holding the bolt and evaluate two robustness metrics: how robustly the weight fixtures the beam and how robustly the robot is able to grasp (and therefore move) the weight to this placement. 
\figref{fig:robustness}-right shows the trade-off: a heavier weight more easily fixtures the beam but is harder to robustly grasp. 
In finding a robust plan, and hence a low cost plan, the planner is incentivized to act like Goldilocks and pick the weight that best balances this trade-off.

\section{Discussion}
\label{sec:discussion}

Our goal is to solve complex multi-step manipulation problems that involve forceful operations between the robot and the environment, where forceful operations are defined as the robot applying a wrench at a pose. 
We leverage an existing task and motion planning system, augmenting it to reason over controllers that exert that wrench, maintain the forceful kinematic chain, and fixture the object we are acting on. 
We demonstrate our system in two example domains: opening a push-and-twist childproof bottle and twisting a nut on a bolt.
%We introduce several actions that, given different ways to apply force and fixture, all use a similar set of underlying kinematics and mechanics tools. 
We also demonstrate the use  of cost-sensitive planning to prioritize actions that are robust to perturbations in the parameters of our stability metrics. 

The current system has several limitations, notably that the the compliant controllers have fixed compliance, the computational cost is high, and the planner generates a fixed sequence of actions rather than a policy, e.g. for turning a nut until it is tight.  These are fertile ground for future work.

\section*{ACKNOWLEDGMENTS}
This work was supported by the NSF Graduate Research Fellowship.
We thank the members of the MCube Lab and LIS Group for their helpful feedback and especially Caelan Garrett for the many PDDLStream discussions.

%\addtolength{\textheight}{-12cm}   
% This command serves to balance the column lengths on the last page of the document manually. It shortens
% the textheight of the last page by a suitable amount. This command does not take effect until the next page
% so it should come on the page before the last. Make sure that you do not shorten the textheight too much.

\newpage
{\footnotesize
\bibliographystyle{ieeetr}
\bibliography{references}}

\end{document}